\documentclass[10pt,twocolumn,letterpaper]{article}
\usepackage[pagenumbers]{cvpr} 

\usepackage[dvipsnames]{xcolor}

\usepackage{makecell}

\definecolor{cvprblue}{rgb}{0.21,0.49,0.74}
\usepackage[pagebackref,breaklinks,colorlinks,citecolor=cvprblue]{hyperref}
\usepackage{algorithmic}
\usepackage[linesnumbered,ruled,vlined]{algorithm2e}
\usepackage[toc,page]{appendix}
\usepackage{multirow}
\usepackage{pifont}
\newcommand{\cmark}{\ding{51}}
\newcommand{\xmark}{\ding{55}}
\newcommand{\tabincell}[2]{\begin{tabular}{@{}#1@{}}#2\end{tabular}}
\newcommand*{\affaddr}[1]{#1} 

\newcommand*{\email}[1]{\texttt{#1}}

\title{Enhancing Visual Document Understanding with Contrastive Learning in \\Large Visual-Language Models}

\author{
	Xin Li\textsuperscript{\dag\thanks{Equal contribution. \textsuperscript{\dag}Contact person.}} \quad Yunfei Wu\textsuperscript{*} \quad Xinghua Jiang \quad Zhihao Guo \quad Mingming Gong \quad Haoyu Cao \quad \\
	Yinsong Liu \quad Deqiang Jiang \quad Xing Sun \\		
	\affaddr{Tencent YouTu Lab} \quad\\
	\email{\small \{fujikoli, marcowu, clarkjiang, nicholasguo, riemanngong, rechycao,}\\
	\email{\small jasonysliu, dqiangjiang, winfredsun\}@tencent.com}
}

\begin{document}
\maketitle
\begin{abstract}
\vspace{-2mm}
Recently, the advent of Large Visual-Language Models~(LVLMs) has received increasing attention across various domains, particularly in the field of visual document understanding~(VDU). 
Different from conventional vision-language tasks, VDU is specifically concerned with text-rich scenarios containing abundant document elements.
Nevertheless, the importance of fine-grained features remains largely unexplored within the community of LVLMs, leading to suboptimal performance in text-rich scenarios.
In this paper, we abbreviate it as the fine-grained feature collapse issue.
With the aim of filling this gap, we propose a contrastive learning framework, termed \textbf{D}ocument \textbf{O}bject \textbf{CO}ntrastive learning~(DoCo), specifically tailored for the downstream tasks of VDU. 
DoCo leverages an auxiliary multimodal encoder to obtain the features of document objects and align them to the visual features generated by the vision encoder of LVLM, which enhances visual representation in text-rich scenarios.
It can represent that the contrastive learning between the visual holistic representations and the multimodal fine-grained features of document objects can assist the vision encoder in acquiring more effective visual cues, thereby enhancing the comprehension of text-rich documents in LVLMs.
We also demonstrate that the proposed DoCo serves as a plug-and-play pre-training method, which can be employed in the pre-training of various LVLMs without inducing any increase in computational complexity during the inference process.
Extensive experimental results on multiple benchmarks of VDU reveal that LVLMs equipped with our proposed DoCo can achieve superior performance and mitigate the gap between VDU and generic vision-language tasks.
\end{abstract}    
\vspace{-6mm}
\section{Introduction}
\begin{figure}[htb]
	\centering
	\includegraphics[width=1\linewidth]{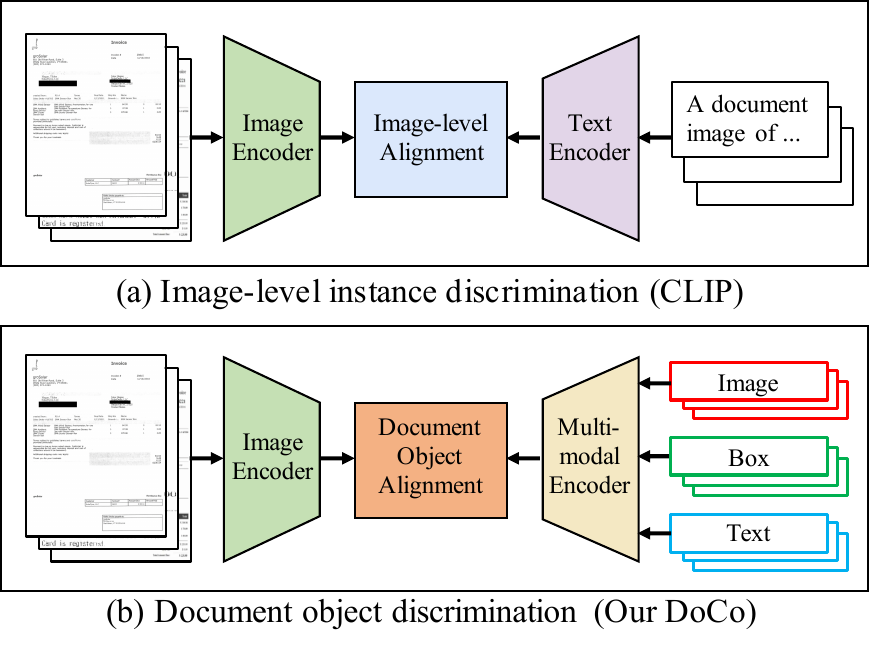}
	\caption{{The motivation of the proposed DoCo.~(a) Image-level instance discrimination between visual and textual inputs, which aims to learn the holistic representations but fails to extract fine-grained features in text-rich scenarios. ~(b) Document object discrimination between visual and multimodal inputs, which enhances the visual representation of image encoder in LVLMs and achieves the better visual understanding performance for VDU.}} 
	\label{fig:motivation}
\end{figure}

Large Language Models~(LLMs)~\cite{brown2020language,scao2022bloom,touvron2302llama,chowdhery2022palm,anil2023palm} have aroused increasing research interest due to their powerful text generation and understanding capabilities. 
These models can be aligned with user intent through fine-tuning instructions, demonstrating strong interactive capabilities and potential to serve as productivity-boosting intelligent assistants. 
However, the inherent limitation of these models is their confinement to the domain of textual data, rendering them incapable of processing other prevalent modalities such as images, speech, and video.
This significantly curtails the applicability of these models.
To circumvent this constraint, a set of Large Visual-Language Models~(LVLMs)~\cite{li2023blip,bai2023qwen,ye2023m,zhu2023minigpt,liu2023visual} has been devised to enhance the ability of LLMs to perceive and understand visual signals. 
These LVLMs have shown immense potential in addressing real-world visual challenges.

Despite these multimodal large models having commendable capabilities in multimodal understanding, their proficiency in text-rich scenarios, such as visual document understanding (VDU) tasks, remains limited due to insufficient specific training.
To address this issue, recent advancements in LVLMs~\cite{ye2023mplug,zhang2023llavar,feng2023unidoc, ye2023ureader} have further explored the performance of VDU in the community, which utilize large-scale document data for training or combine multiple VDU tasks of multimodal understanding.
For example, mPLUG-DocOwl~\cite{ye2023mplug} constructs a comprehensive dataset dedicated to document image comprehension. 
LLaVAR~\cite{zhang2023llavar} proposes incorporating a text recognition pre-training task to enhance the understanding of text-rich images. 
UniDoc~\cite{feng2023unidoc} implements a unified multimodal instruction tuning on the contributed large-scale instruction following datasets. 
While these works have made significant strides in text-rich scenarios, they still struggle with accurately capturing the textual details within images. 
As depicted in Fig.~\ref{fig:motivation}(a), the vision encoders of these models are initialized from CLIP-based~\cite{radford2021learning} weights, which are pre-trained through contrastive learning to foster cross-modality alignment between visual and textual inputs via image-level instance discrimination.
Nevertheless, these image-level representations are not optimal for dense prediction tasks such as VDU and fail to extract fine-grained visual features.
This dilemma leads to the following question: \textit{how can the exploration of fine-grained features enhance the performance of LVLMs  in text-rich scenarios?}
We define it as the fine-grained collapse issue, which still lacks investigation in the VDU field.

In this work, we introduce a novel \textbf{D}ocument \textbf{O}bject \textbf{CO}ntrastive learning~(DoCo) tailored for this problem, as illustrated in Fig.~\ref{fig:motivation}(b).
Concretely, we adopt a contrastive learning approach, which leverages a multimodal encoder to obtain the multimodal features~(\textit{i.e.}, visual, layout and textual) of document objects and align them to the visual features produced by the image encoder of LVLM.
During the pre-training phase, the image encoder processes the whole image into a series of embeddings consistent with the visual feature extraction of LVLMs.
Beyond visual features, the layout information along with the textual contents of the document objects derived from the OCR engine is embedded to the multimodal encoder to capture the fine-grained features.
Subsequently, the output representations generated by the two encoders are trained to align in the manner of contractive learning.
In this way, the image encoder of LVLMs can acquire more effective visual cues from the document objects, thereby strengthening the understanding ability of text-rich scenarios.
Note that the yielded DoCo is a plug-and-play pre-training method, which can be applied in the pre-training of various LVLMs without inducing any increase in computational complexity during inference.
To sum up, our main contributions are in the three folds:
\begin{itemize}
	\item We investigate the importance of fine-grained features in VDU tasks and propose the fine-grained feature collapse issue. To our best knowledge, we are the first to research the enhancement of visual information for LVLMs.
	\item We coin a novel DoCo tailored for the fine-grained feature collapse issue, which introduces a contrastive learning framework to assist LVLMs in acquiring more beneficial visual features in text-rich scenarios and enhancing the perception of document objects.
	\item Experimental results on various VDU datasets demonstrate that our proposed method facilitates significant performance enhancements for LVLMs in text-rich scenes.
\end{itemize}
\section{Related Work}
\begin{figure*}[t]
	\centering
	\includegraphics[width=1\linewidth]{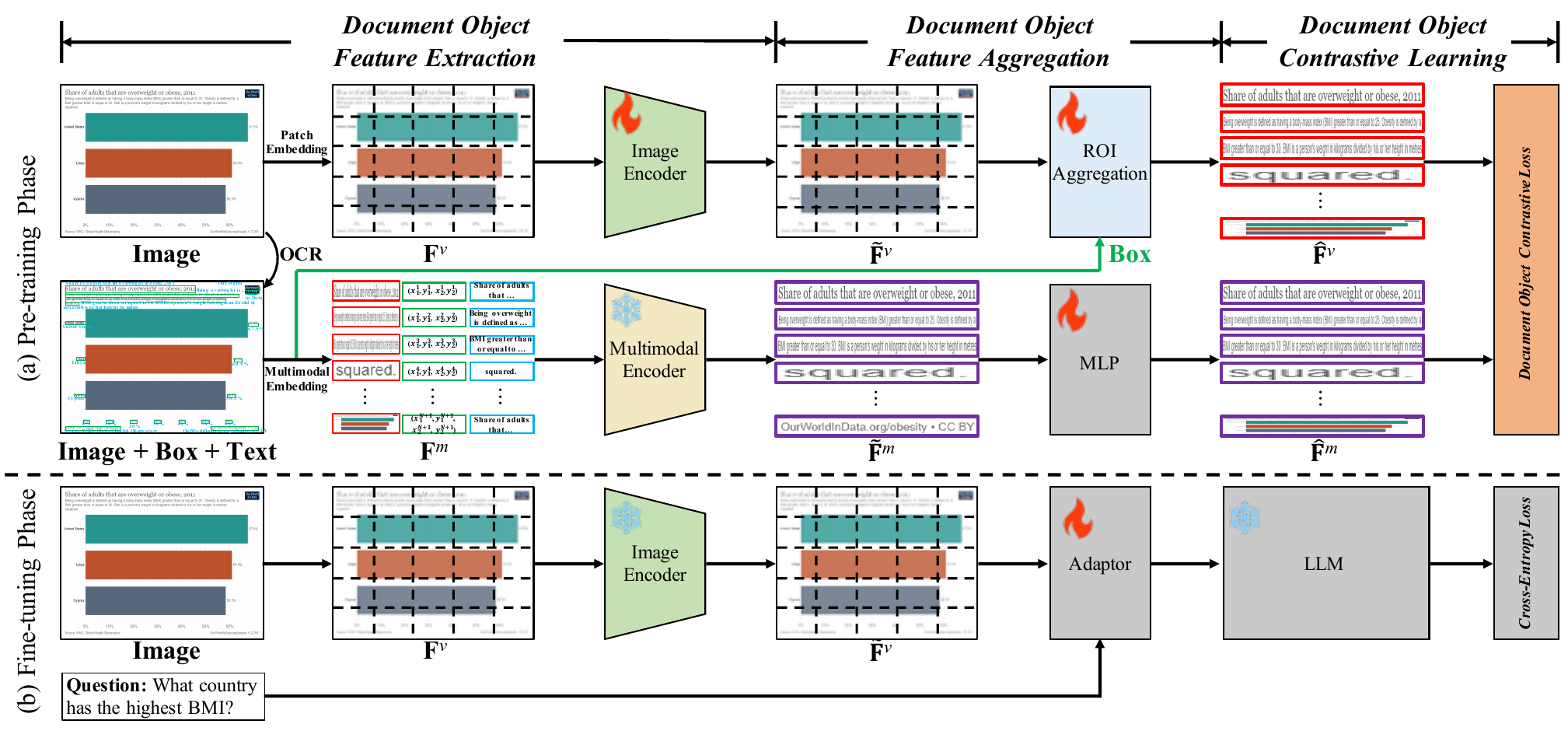}
	\vspace{-5mm}
	\caption{A schematic overview of our proposed DoCo. We aim to bridge the representation learning gap between visual features and multimodal features by guiding the former to imitate the features extracted by the latter. Note the branch of multimodal feature extraction is removed after pre-training, which indicates the computational complexity does not increase during the phase of fine-tuning and inference.} 
	\label{fig:architecture}
	\vspace{-2mm}
\end{figure*}

\subsection{Visual Document Understanding}
The field of Visual Document Understanding~(VDU) is concerned with the interpretation and comprehension of a wide array of digital-native or scanned documents, which includes but is not limited to forms, tables, reports, and academic research papers.
The techniques employed in VDU can be broadly classified into two primary categories.
The initial category aims to address the task by integrating images and OCR annotations sourced from external OCR systems~\cite{xu2020layoutlm, hong2022bros, xu2020layoutlmv2, huang2022layoutlmv3, tang2023unifying}.
Among these, LayoutLMv3~\cite{huang2022layoutlmv3} utilizes a streamlined transformer-based architecture and loss functions that foster alignment between the patches and OCR.
UDOP~\cite{tang2023unifying} employs a unified encoder to represent features from both images and texts, thereby transforming information from these modalities into vision-text embeddings.
The second category investigates methods that directly process the document images without external OCR tools.
Donut~\cite{kim2021donut} proposes a model that performs text reading from document images as a pre-training task without the OCR engine involved.
Pix2Struct~\cite{lee2023pix2struct} introduces the concept of screenshot parsing objectives, while MATCHA~\cite{liu2022matcha} incorporates chart derendering and math reasoning into the model training.

\subsection{Large Visual-Language Models}
Large Language Models~(LLMs), such as GPT-3~\cite{brown2020language}, PaLM~\cite{chowdhery2022palm}, PaLM2~\cite{anil2023palm}, BLOOM~\cite{scao2022bloom}, and LLaMA~\cite{touvron2302llama}, have showcased remarkable zero-shot capabilities across a wide range of open-ended tasks.
In recent years, researchers have been exploring the integration of high-performance LLMs with Large Vision-Language Models~(LVLMs) to enable a deeper understanding of visual inputs.
BLIP-2~\cite{li2023blip} aligns queried visual features with text using multiple vision-language losses and employs a simple fully connected layer to feed the queried embedding into a frozen language model.
Mini-GPT4~\cite{zhu2023minigpt}, Qwen-VL~\cite{bai2023qwen} and mPLUG-Owl~\cite{ye2023m} retain the Q-Former architecture while replacing the language model with a larger one.
These models are then fine-tuned on meticulously collected instruction data to enhance their performance.
To extend the capabilities of LVLMs to VDU, mPLUG-DocOwl~\cite{ye2023mplug} proposes a modularized model based on mPLUG-Owl for OCR-free document understanding.
UniDoc~\cite{feng2023unidoc} employs a universal large multimodal model to simultaneously perform text detection, recognition, spotting, and understanding.
UReader~\cite{ye2023ureader} designs a shape-adaptive cropping module before the encoder-decoder architecture to leverage the frozen low-resolution vision encoder for processing high-resolution images.
These comprehensive methods underscore the potential of LVLMs in tackling intricate vision-language tasks.
However, the significance of fine-grained features, which are crucial for VDU, remains unexplored.

\subsection{Contrastive Learning}
\par \setlength{\parindent}{0em}
Contrastive learning methods have been extensively employed to derive meaningful representations, demonstrating their efficacy across a multitude of tasks such as image classification~\cite{radford2021learning, khosla2020supervised}, object detection~\cite{xie2021detco, sun2021fsce}, and image segmentation~\cite{zhao2021contrastive, chaitanya2020contrastive}.
The vision-language framework, CLIP~\cite{radford2021learning}, which is extensively adopted, capitalizes on contrastive learning and vast data quantities to align images and texts in a semantic space, resulting in strong generalization capabilities across a wide range of downstream tasks.
Nevertheless, the discrimination of the entire image between the visual and textual information is not favorable for the comprehension of visual documents due to the absence of fine-grained features.
To address this issue, we introduce a novel contrastive feature enhancement technique, which effectively augments the visual features within LVLMs, resulting in a robust capability of analyzing text-rich images. 

\section{Methodology}

\subsection{Overall Architecture}
The overview of the proposed \textbf{D}ocument \textbf{O}bject \textbf{CO}ntrastive learning~(DoCo) can be observed in Fig.~\ref{fig:architecture}, whose goal is to align the multimodal features of document objects to the image representations.
Here, we employ various LVLMs~(\textit{i.e.}, Qwen-VL-Chat~\cite{bai2023qwen} and mPLUG-Owl~\cite{ye2023m}) to illustrate our fundamental design principle.
DoCo mainly comprises two branches, the vision encoder, situated at the top, and the multimodal encoder, positioned at the bottom.
During the pre-training phase, an input image undergoes processing by the visual encoder to obtain a series of image embeddings.
Concurrently, an Optical Character Recognition~(OCR) engine is employed to parse the bounding boxes and contents of texts, which are subsequently sent to the multimodal encoder for the extraction of the fused features of document texts.
Furthermore, the output representations generated by both encoders are aligned at the document object level by the ROI Aggregation module.
Ultimately, the visual representations from the image encoder are enhanced by DoCo within the contrastive learning paradigm, encompassing both Intra-Document Object Contrastive Learning~(Intra-DoCo) and Inter-Document Object Contrastive Learning~(Inter-DoCo).
During the fine-tuning phase, the multimodal feature extraction branch is removed, while the enhanced image features, in conjunction with the embeddings from additional text input~(\textit{e.g.}, question), are fed into the vision-language adapter and subsequently into the LLM.

\subsection{Document Object Feature Extraction}
\textbf{Visual features:} 
Throughout the training and inference phases, the input image $\mathbf{x}$ is transformed by the process of patch embedding into a sequence of image embeddings denoted as $\mathbf{F}^{v}$.
Subsequent to this procedure, the conventional Vision Transformer~(ViT)~\cite{dosovitskiy2020image} initialized with the pre-trained weights derived from CLIP~\cite{radford2021learning} is employed to extract the embeddings to a collection of visual features $\mathbf{\widetilde{F}}^{v} = \left\lbrace \mathbf{f}^{v}_{1, 1}, \mathbf{f}^{v}_{1, 2}, ..., \mathbf{f}^{v}_{H',W'} \right\rbrace \in \mathbb{R}^{(H'W') \times d_{v}}$ are produced, where $(H'W')$ denotes the patch size of the image and $d_{v}$ is the dimension of patch.

\textbf{Multimodal features:} 
In this process, we aim to extract the multimodal representations of document objects, which amalgamates various embeddings to facilitate the feature extraction of each box via a multimodal encoder.
Drawing inspiration from recent pre-training frameworks for document understanding, we adhere to  LayoutLMv3~\cite{huang2022layoutlmv3} to leverage the multimodal features $\mathbf{{F}}^{m}$ consisting of visual embeddings, layout embeddings and textual embeddings, as illustrated by the red, green and blue boxes in Fig.~\ref{fig:architecture}.
Following this procedure, a set of multimodal embeddings $\mathbf{\widetilde{F}}^{m} = \left\lbrace \mathbf{f}^{m}_{1},  \mathbf{f}^{m}_{2}, ..., \mathbf{f}^{m}_{N+1}\right\rbrace \in \mathbb{R}^{(N + 1) \times d_{m}}$ are generated by the multimodal encoder, where $N$ signifies the number of boxes recognized by the OCR engine and $d_{m}$ means the dimension of the aggregated features of the objects.
Specially, a box possessing an identical resolution to the image is utilized to extract the global features, thereby resulting in the quantity of the multimodal embeddings being $N+1$.
A comprehensive description of the multimodal feature extraction is provided in the supplementary material.

\begin{figure}[htp]
	\centering
	\includegraphics[width=1\linewidth]{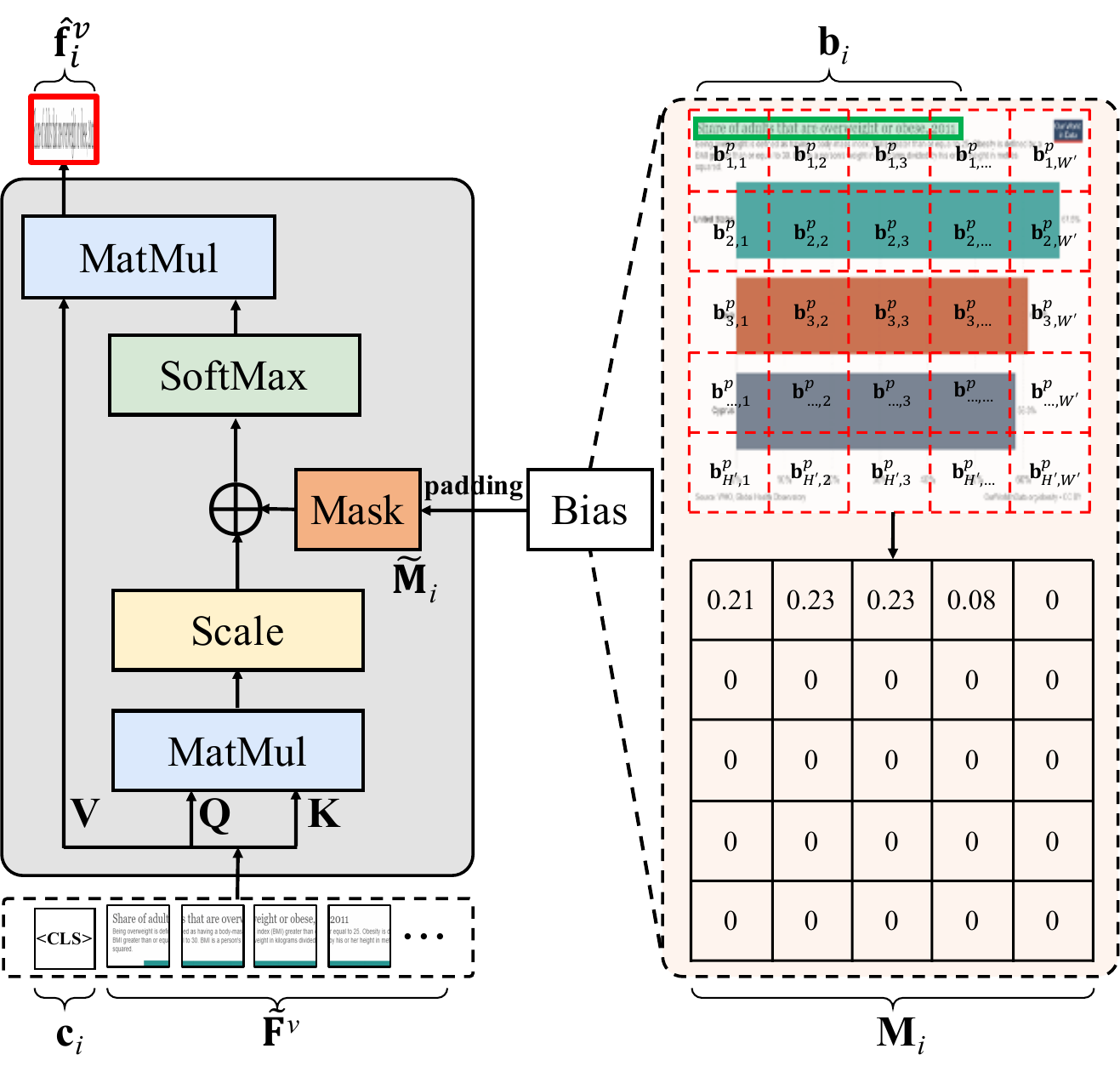}
	\vspace{-4mm}
	\caption{The proposed ROI Aggregation. The dashed red grid represents the image patch features, and the solid green region denotes the bounding box region. The overlap between each patch and the given region is calculated and serves as the attention mask for the visual aggregation of the document objects. Best viewed in color.} 
	\label{fig:alignment}
	\vspace{-2mm}
\end{figure}
\subsection{Document Object Feature Aggregation}
Upon the completion of the visual encoding phase, we attempt to accurately acquire the image features that correspond to a specified region of the document object.
Nevertheless, the direct application of a traditional feature extraction technique, such as ROIAlign~\cite{he2017mask} or ROIPooling~\cite{girshick2015fast}, to a ViT backbone presents a challenge due to the disparate shapes of the output feature maps produced by CNN and ViT.
Given that the bounding box coordinates are expressed in pixel units, the bounding box edges frequently fail to align with the patch boundaries.
To address this issue, we propose a novel \textit{ROI Aggregation} module for the feature extraction, which leverages the overlapped area between each patch and the specified region to the attention mask in the attention function.

\textbf{ROI Aggregation:} 
As depicted in Fig.~\ref{fig:alignment}, the patch features output by the image encoder are denoted as $\mathbf{\widetilde{F}}^{v}$.
For the $i$-th box object, the corresponding bounding box is defined as $\mathbf{b}_{i} = \left\lbrace x_{1}^{i}, y_{1}^{i}, x_{2}^{i}, y_{2}^{i} \right\rbrace$.
Initially, we calculate the overlapped area between the specified region $\mathbf{b}_{i}$ and each patch $\mathbf{b}^{p} = \left\lbrace \mathbf{b}^{p}_{1, 1}, \mathbf{b}^{p}_{1, 2}, ..., \mathbf{b}^{p}_{H', W'} \right\rbrace  \in \mathbb{R}^{(H'W') \times 4}$, and normalize it to $\mathbf{M}_{i}$ with the patch area:
\begin{equation}
	\small
	\begin{aligned}
		\mathbf{M}_{i} = Overlap(\mathbf{b}_{i}, \mathbf{b}^{p}) / Area(\mathbf{b}^{p}),
	\end{aligned}
\end{equation}
where $Overlap$ and $Area$ denote calculating the overlap area between two boxes and the area of the box, respectively.
$\mathbf{M}_{i} \in \mathbb{R}^{(H'W') \times 1}$ represents the attention bias of the visual features towards the $i$-th document object.
Subsequently, we prepend a $\textless {\rm CLS} \textgreater$ token $\mathbf{c}_{i}$ to the sequence of patch embeddings $\mathbf{\widetilde{F}}^{v}$, denoted as the reorganized sequence $\left\lbrace\mathbf{c}_{i}, \mathbf{\widetilde{F}}^{v}\right\rbrace  \in \mathbb{R}^{(H'W' + 1) \times d_{v}}$, to aggregate the image representations of the document box.
Note that $\mathbf{M}_{i}$ is then padding with zeros to $\mathbf{\widetilde{M}}_{i}  \in \mathbb{R}^{(H'W' + 1) \times (H'W' + 1)}$, which functions as the attention bias for the attention score.
Ultimately, the reorganized sequence regarded as $\mathbf{Q}$, $\mathbf{K}$, and $\mathbf{V}$, along with $\mathbf{\widetilde{M}}_{i}$, is process by the self-attention layer:
\begin{equation}
	\small
	\begin{aligned}
	Attention(\mathbf{Q,K,V}) = softmax(\frac{\mathbf{Q}\cdot\mathbf{K}^\top}{\sqrt{d_v}} + \mathbf{\widetilde{M}}_{i}) \cdot \mathbf{V},
	\end{aligned}
\end{equation}
Following this procedure, the aggregated image features of the $i$-th box object are obtained, denoted as $\mathbf{\hat{f}}^{v}_{i}$, and a set of visual embeddings of the document boxes $\mathbf{\hat F}^{v} = \left\lbrace \mathbf{\hat{f}}^{v}_{1}, \mathbf{\hat{f}}^{v}_{1}, ..., \mathbf{\hat{f}}^{v}_{N+1} \right\rbrace \in \mathbb{R}^{(N+1) \times d_{v}}$ are generated.
Furthermore, two groups of MLP layers are employed to project the multimodal representations into a uniform dimensional space, thereby converting the dimension $d_{m}$ of $\mathbf{\widetilde{F}}^{m}$ to the dimension $d_{v}$ of $\mathbf{\hat {F}}^{m} = \left\lbrace \mathbf{\hat{f}}^{m}_{1}, \mathbf{\hat{f}}^{m}_{1}, ..., \mathbf{\hat{f}}^{m}_{N+1} \right\rbrace \in \mathbb{R}^{(N+1) \times d_{v}}$.

\subsection{Document Object Contrastive Learning}
In an effort to incorporate the visual and multimodal features of document objects into a shared domain, we present a novel DoCo designed to improve the visual representations, which significantly contributes to the extraction of the visual features in text-rich scenes.
The proposed scheme consists of two components: Intra-DoCo and Inter-DoCo, which respectively represent the learning of document object representation within a single image and between two distinct images.
\begin{figure}[h]
	\centering
	\includegraphics[width=1\linewidth]{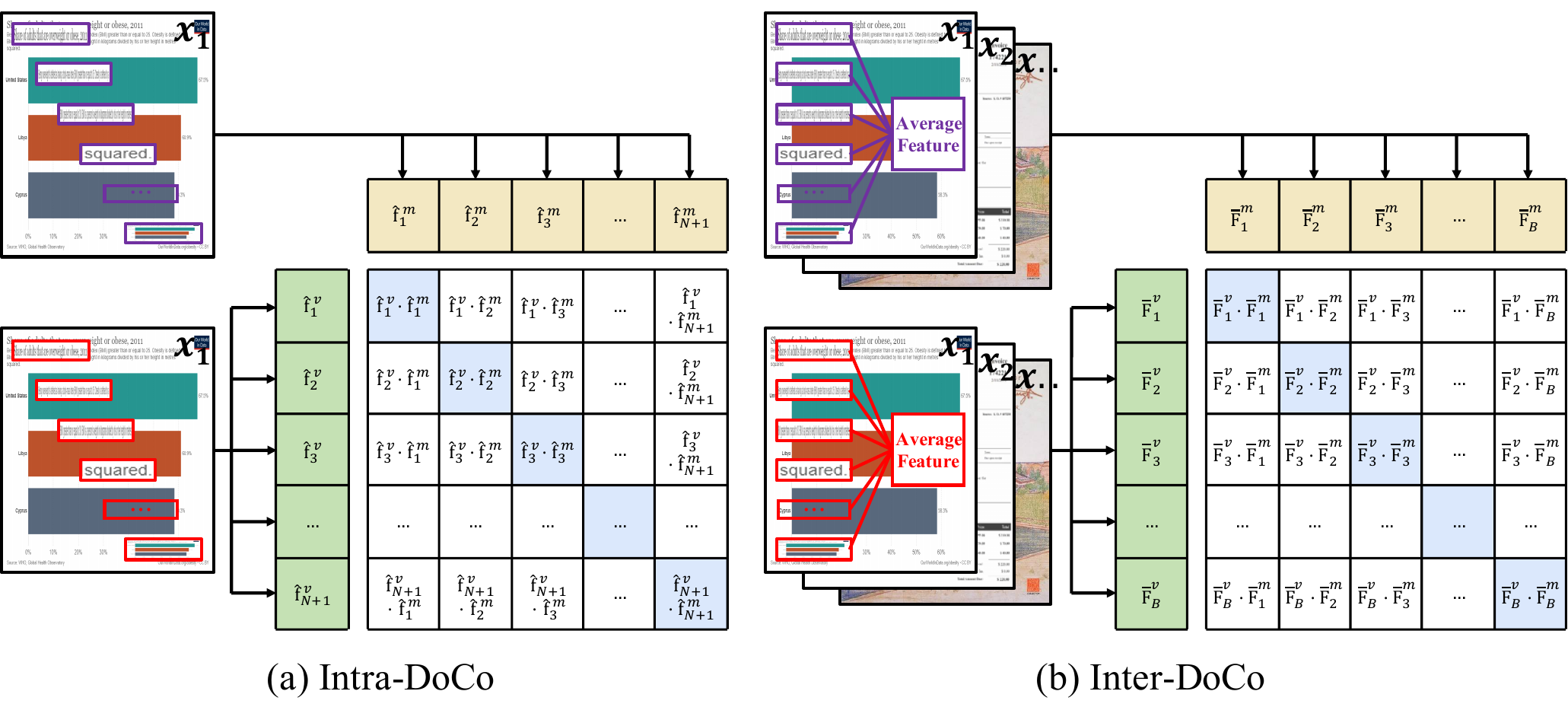}
	\caption{An illustrative view of DoCo. The red and purple boxes represent the aggregated visual and multimodal features of document objects, respectively. Best viewed in color.} 
	\label{fig:cl}
\end{figure}

\textbf{Intra-DoCo:} 
For the purpose of achieving more fine-grained representations in text-rich scenes, we establish an alignment between visual and multimodal features at the level of document object within the image, as opposed to discriminating image and text features at the overall image level.
As depicted in Fig.~\ref{fig:cl} (a), the visual and multimodal features of boxes within an image are denoted as 
$\mathcal{F}^{v}_{intra} = \left\lbrace {\mathbf{\hat{f}}}^{v}_{1}, {\mathbf{\hat{f}}}^{v}_{2}, ..., {\mathbf{\hat{f}}}^{v}_{N+1} \right\rbrace \in \mathbb{R}^{(N+1)  \times d_{v}}$ and $\mathcal{F}^{m}_{intra} = \left\lbrace {\mathbf{\hat{f}}}^{m}_{1}, {\mathbf{\hat{f}}}^{m}_{2}, ..., {\mathbf{\hat{f}}}^{m}_{N+1} \right\rbrace \in \mathbb{R}^{(N+1) \times d_{v}}$ respectively.
The objective of Intra-DoCo is to predict the actual pairings from the $(N+1)$ × $(N+1)$ possible ($\mathcal{F}^{v}_{intra}$, $\mathcal{F}^{m}_{intra}$) combinations across a batch, which aims to maximize the cosine similarity of the visual and multimodal embeddings of the $N + 1$ pairs illustrated by the blue blocks of Fig.~\ref{fig:cl} (a), while simultaneously minimizing the cosine similarity of the embeddings of the $(N+1)^2 - (N+1)$ incorrect pairings shown with the white blocks in Fig.~\ref{fig:cl} (a).
Given that the $i$-th visual representations are $\mathbf{\hat{f}}^{v}_{i} \in \mathcal{F}^{v}_{intra}$, the corresponding positive multimodal features are $\mathbf{\hat{f}}^{m}_{i} \in \mathcal{F}^{m}_{intra}$, and the negative multimodal features are $\mathbf{\hat{f}}^{m}_{j} \in \mathcal{F}^{m}_{intra}$ and $i \neq j$, the optimization is then defined as the sum of minimizing the similarity of positive sample pairs and maximizing the similarity of negative sample pairs:
\begin{equation}
	\small
	\begin{aligned}
		\mathcal{L}_{\text{Intra-DoCo}}^{\textbf{x}} = - \frac{1}{N+1}\sum_{i = 1}^{N+1} log \left( \frac{{e^{{sim}(\mathbf{\hat{f}}^{v}_{i}, \mathbf{\hat{f}}^{m}_{i})}}}{\sum_{j = 1}^{N+1}{e^{{sim}(\mathbf{\hat{f}}^{v}_{i}, \mathbf{\hat{f}}^{m}_{j})}}} \right),
	\end{aligned}
\end{equation}
where $sim$ computes similarity scores between the pairs.
In a symmetrical manner, we also calculate the loss from $j$ to $i$ as $ \mathcal{L}_{\text{Intra-DoCo}}^{\textbf{y}}$ in accordance with the CLIP model. 
And Intra-DoCo is optimizd using the InfoNCE loss function:
\begin{equation}
	\small
	\begin{aligned}
		\mathcal{L}_{\text{Intra-DoCo}} = (\mathcal{L}_{\text{Intra-DoCo}}^{\textbf{x}} + \mathcal{L}_{\text{Intra-DoCo}}^{\textbf{y}}) / 2.
	\end{aligned}
\end{equation}
\textbf{Inter-DoCo:} 
In contrast to the conventional approach of learning contrastive representations with the image-level instance discrimination, Inter-DoCo emphasizes the comprehensive perspective of document objects across diverse images.
As illustrated in Fig.~\ref{fig:cl} (b),
given a batch of $B$ images $\mathbf{X} = \left\lbrace x_{1}, x_{2}, ..., x_{B} \right\rbrace \in \mathbb{R}^{B \times H \times W \times C}$, the extracted visual and multimodal features are represented as ${\mathbb{\hat{F}}}^{v} = \left\lbrace \mathbf{\hat F}^{v}_{1}, \mathbf{\hat F}^{v}_{2}, ..., \mathbf{\hat F}^{v}_{B} \right\rbrace \in \mathbb{R}^{B \times (N+1) \times d_{v}}$ and ${\mathbb{\hat{F}}}^{m} = \left\lbrace \mathbf{\hat F}^{m}_{1}, \mathbf{\hat F}^{m}_{2}, ..., \mathbf{\hat F}^{m}_{B} \right\rbrace \in \mathbb{R}^{B \times (N+1) \times d_{v}}$ respectively. 
To derive the global features of document objects, these visual and multimodal features are averaged to $\mathcal{F}^{v}_{inter} = \left\lbrace \mathbf{\overline F}^{v}_{1}, \mathbf{\overline F}^{v}_{2}, ..., \mathbf{\overline F}^{v}_{B} \right\rbrace \in \mathbb{R}^{B  \times d_{v}}$ and $\mathcal{F}^{m}_{inter} = \left\lbrace \mathbf{\overline F}^{m}_{1}, \mathbf{\overline F}^{m}_{2}, ..., \mathbf{\overline F}^{m}_{B} \right\rbrace \in \mathbb{R}^{B \times d_{v}}$ respectively. 
Similarly, the computation from $i$ to $j$ is defined as follows:

\begin{equation}
	\small
	\begin{aligned}
	\mathcal{L}_{\text{Inter-DoCo}}^{\textbf{x}} = - \frac{1}{B}\sum_{i = 1}^{B} log \left( \frac{{e^{{sim}(\mathbf{\overline F}^{v}_{i}, \mathbf{\overline F}^{m}_{i})}}}{\sum_{j = 1}^{B}{e^{{sim}(\mathbf{\overline F}^{v}_{i}, \mathbf{\overline F}^{m}_{j})}}} \right),
	\end{aligned}
\end{equation}
and Inter-DoCo is defined as:
\begin{equation}
	\small
	\begin{aligned}
	\mathcal{L}_{\text{Inter-DoCo}} = (\mathcal{L}_{\text{Inter-DoCo}}^{\textbf{x}} + \mathcal{L}_{\text{Inter-DoCo}}^{\textbf{y}}) / 2.
	\end{aligned}
\end{equation}
Considering both Intra-DoCo and Inter-DoCo, the total loss of DoCo is formulated as:
\begin{equation}
	\small
	\begin{aligned}
	\mathcal{L}_{\text{DoCo}} = \mathcal{L}_{\text{Intra-DoCo}} + \mathcal{L}_{\text{Inter-DoCo}}.
	\end{aligned}
\end{equation}

\subsection{Training Strategies}
\textbf{Pre-training:} 
During the pre-training phase, the multimodal backbone remains static, while the image encoder undergoes optimization via the DoCo loss function to enhance the visual feature representations.

\textbf{Fine-tuning:}
Subsequent to the pre-training phase, the weights of the enhanced image encoder are utilized for fine-tuning, and the multimodal encoder is discarded.
We keep the image encoder and LLM frozen during fine-tuning, and update the parameters of the adaptor.
Note we experiment with various LVLMs, the adaptor represents the general module for vision and language feature alignment, that is Position-Aware Vision-Language Adapter for Qwen-VL-Chat~\cite{bai2023qwen} and Visual Abstractor for mPLUG-Owl~\cite{ye2023m}.
The primary aim of the training process is to minimize the cross-entropy of the textual tokens.
\section{Experiments}
\subsection{Datasets} \label{sec:datasets}
\textbf{Pre-training datasets:}
In order to guarantee the richness of text within the pre-training datasets, we adhere to the methodologies proposed by LLaVA~\cite{liu2023visual} and LLaVAR~\cite{zhang2023llavar} for the construction of these datasets.
Specifically, these datasets are comprised of approximately 1.0 million image-text pairs.
This includes a subset of 0.6 million data from the CC3M dataset~\cite{sharma2018conceptual}, utilized by LLaVA~\cite{liu2023visual} during the pre-training phase, and an additional subset of 0.4 million data from the LAION dataset~\cite{schuhmann2022laion}, employed by LLaVAR~\cite{zhang2023llavar} in the same phase.
It is important to note that these aggregated datasets undergo processing with the OCR tool PaddleOCR~\cite{du2020pp} to extract bounding boxes and textual contents, thereby optimizing our proposed DoCo model.

\begin{table}[h]
	\small
	\centering
	\setlength{\tabcolsep}{1.6mm}
	\begin{tabular}{l|ccc}
		\toprule[1.5pt]
		Dataset & Task  &  \tabincell{c}{Fine-tuning \\(Amt)}& \tabincell{c}{Evaluation \\ (Amt)} \\
		\hline
		TextVQA~\cite{singh2019textvqa} & VQA  & 35K &  5.7k \\
		\hline
		DocVQA~\cite{mathew2021docvqa} & VQA & 39K & 5.2k \\ 
		\hline
		ChartQA~\cite{masry2022chartqa} & VQA & 28K & 2.5k \\ 
		\hline
		OCRVQA$^*$~\cite{mishra2019ocr} & VQA & 200K & 100k \\ 
		\hline
		InfoVQA~\cite{mathew2022infographicvqa} & VQA & 24K& 3.3k  \\ 
		\hline
		KLC~\cite{stanislawek2021kleister} & KIE & 14k & 4.9k \\
		\hline
		WTQ~\cite{pasupat2015wtq} & Table & 14k & 4.3k\\
		\hline
		TextCaps~\cite{sidorov2020textcaps} & Image Caption & 91K  & 16k\\ 
		\bottomrule[1.5pt]
	\end{tabular}
	\caption{Statistics of the fine-tuning and evaluation datasets. ``OCRVQA$^*$'' denotes that we sample 25\% data from the original dataset. ``Amt'' means ``Amount''. }
	\label{tab:training_dataset}
\end{table}

\textbf{Fine-tuning datasets:}
During the fine-tuning phase, LVLMs undergo training utilizing an array of text-rich datasets, which include TextVQA~\cite{singh2019textvqa}, DocVQA~\cite{mathew2021docvqa}, ChartQA~\cite{masry2022chartqa}, OCRVQA~\cite{mishra2019ocr}, InfoVQA~\cite{mathew2022infographicvqa}, KLC~\cite{stanislawek2021kleister}, WTQ~\cite{pasupat2015wtq} and TextCaps~\cite{sidorov2020textcaps}.
In Tab.~\ref{tab:training_dataset}, we provide a summary of these datasets, which collectively comprise approximately 0.4 million fine-tuning data.

\textbf{Evaluation datasets:}
We perform extensive experiments on the benchmark datasets reported in Tab.~\ref{tab:training_dataset}.
In alignment with previous studies, the performance on DocVQA~\cite{mathew2021docvqa} and InfoVQA~\cite{mathew2022infographicvqa} is evaluated by the Average Normalized Levenshtein Similarity (ANLS).
For the KLC~\cite{stanislawek2021kleister} dataset, the F1-score is utilized as the evaluation criterion.
The primary metric for WTQ~\cite{pasupat2015wtq}, OCRVQA~\cite{mishra2019ocr}, and TextVQA~\cite{singh2019textvqa} is accuracy.
In the context of ChartQA~\cite{masry2022chartqa}, a relaxed accuracy metric is employed.
Finally, the TextCaps~\cite{sidorov2020textcaps} dataset is assessed using the Consensus-based Image Description Evaluation (CIDEr) metric.

\subsection{Implementation Details} \label{sec:settings}
The experimental framework utilizes Qwen-VL-Chat~\cite{bai2023qwen} and mPLUG-Owl~\cite{ye2023m} as foundational models to illustrate the competencies of DoCo.
The image encoder utilizes the Vision Transformer~(ViT)~\cite{dosovitskiy2020image} architecture, initialized with the pre-trained weights from the ViT-bigG~\cite{radford2021learning} of Openclip.
For the multimodal encoder, the LayoutLMv3$_{LARGE}$~\cite{huang2022layoutlmv3} is employed to extract the multimodal features of document objects, whose configuration incorporates a 24-layer transformer encoder, 16 self-attention heads, a hidden size of 1,024, and an intermediate size of 4,096 for the feed-forward network.
During the training and inference stages, the images are resized to 448$^{2}$.
All experiments are conducted on a computational platform consisting of 128 NVIDIA A100 80GB GPUs.
The models undergo pre-training with a cumulative batch size of 640 over the course of 1 epoch, followed by fine-tuning with a total batch size of 256 over the span of 5 epochs.
The models are trained utilizing the AdamW optimizer with $\beta_1=0.9$, $\beta_2 = 0.98$, $eps = 1e^{-6}$.
The cosine learning rate schedule is adopted, setting the maximum learning rate at $2e^{-4}$ and minimum at $1e^{-6}$, with a linear warm-up of 500 steps.
The models incorporate a weight decay of $5e^{-2}$ and a gradient clipping of 1.0 to ensure the stability of the training process.

\begin{table*}[t]
	\small
	\setlength{\tabcolsep}{2.2mm}
	\centering
	\begin{tabular}{l|c|cccccccccc}
		\toprule[1.5pt]
		Method & Resolution & OCRVQA & TextVQA & DocVQA & InfoVQA & ChartQA & KLC &WTQ  & TextCaps\\
		\hline
		MiniGPT-4\ddag~\cite{zhu2023minigpt} & 224$^{2}$ &  11.5 & 18.7 & 3.0 & 13.3 & 4.3 & - & - & -\\
		mPLUG-Owl\ddag~\cite{ye2023m} & 224$^{2}$ &28.6 & 40.2 & 6.9 & 16.5 & 9.5 & - & - & -\\
		Qwen-VL~\cite{bai2023qwen} & 448$^{2}$ & 75.7 & 63.8 & 65.1 & - & 65.7 & - & - & - \\
		Qwen-VL-Chat~\cite{bai2023qwen} & 448$^{2}$ & 70.5 & 61.5 & 62.6 & - & 66.3 & - & - & -\\
		mPLUG-DocOwl~\cite{ye2023mplug} & - & - & 52.6 & 62.2 & 38.2 & 57.4 & 30.3 & 26.9 & 111.9\\
		LLaVAR(336)~\cite{zhang2023llavar} & 336$^{2}$ & 23.8&48.5&11.6&-&- & - & - & - \\
		UReader~\cite{ye2023ureader} & 224$^{2}$ & - & 57.6 & 65.4 & 42.2 & 59.3 & 32.8 & 29.4 & 118.4 \\
		\hline
		Qwen-VL-Chat$^{\dag}$ & 448$^{2}$ & {71.1} & {61.7} & {62.2} & {33.1} & {67.3}  & {31.5}  & {24.8}  & {112.3} \\
		\textbf{Qwen-VL-Chat$^{\dag\dag}$}& 448$^{2}$ & \textbf{73.2} & \textbf{63.6} & \textbf{64.8} & \textbf{34.9} & \textbf{68.9}  & \textbf{33.8}  & \textbf{26.9}  & \textbf{114.5} \\
		mPLUG-Owl$^{\dag}$ & 448$^{2}$ & {70.3} & {53.5} & {61.8} & {32.5} & {58.3}  & {31.2}  & {25.2}  & {113.4}\\
		\textbf{mPLUG-Owl$^{\dag\dag}$} & 448$^{2}$ & \textbf{72.1} & \textbf{55.7} & \textbf{63.6} & \textbf{34.1} & \textbf{60.1}  & \textbf{32.9}  & \textbf{26.4}  & \textbf{115.9}\\
		\bottomrule[1.5pt]
	\end{tabular}
	\caption{Comparison results of different LVLMs on various benchmarks of visual document understanding. We use ``$\ddag$'' to refer to the results fetched from ~\cite{liu2023hidden}. The models with ``$\dag$'' and  ``$\dag\dag$'' denote pre-training with CLIP and DoCo respectively, which are optimized with the same datasets and experimental settings for a fair comparison.}
	\label{tab:model_result}
\end{table*}
\begin{figure*}[t]
	\centering
	\includegraphics[width=1\linewidth]{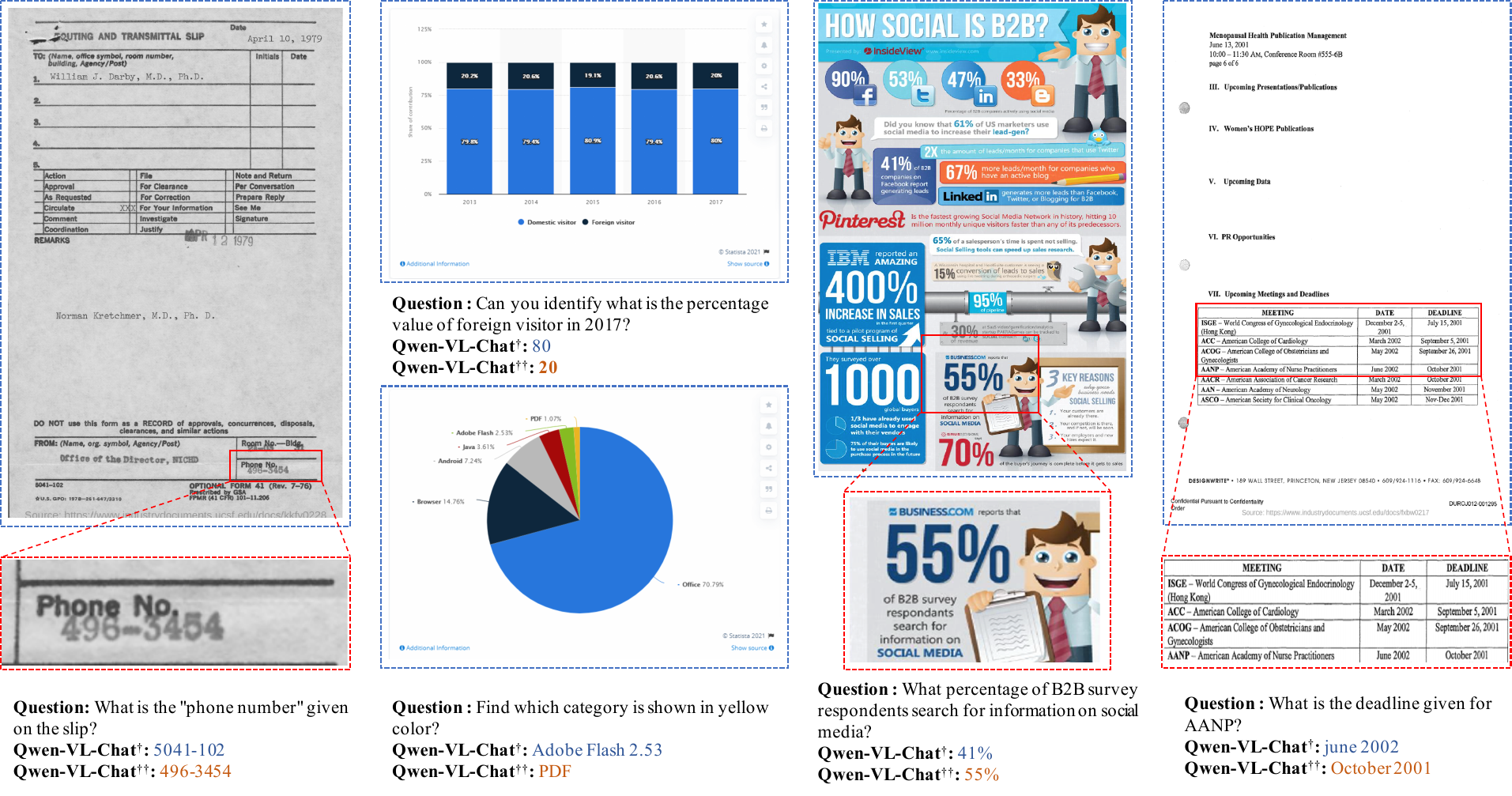}
	\caption{{Qualitative results between CLIP~(``$\dag$'') and DoCo~(``$\dag\dag$''). Crucial regions are enlarged for clearer visualization.}} 
	\label{fig:qualitative_results}
\end{figure*}
\subsection{Comparison with State-of-the-arts}
The performance outcomes of various LVLMs on different benchmarks pertaining to visual document understanding are consolidated in Tab.~\ref{tab:model_result}, from which we can see that DoCo significantly enhances the efficacy of LVLMs.
In detail, the mean score of the Qwen-VL-Chat~\cite{bai2023qwen}, when integrated with DoCo, surpasses the score of the model pre-trained by CLIP~\cite{radford2021learning} by approximately 2\%, which demonstrates the critical role of fine-grained document features.
To further investigate the generalization benefits of the proposed framework, a series of experiments are conducted to juxtapose it with other SOTAs.
When evaluating our DoCo on mPLUG-Owl~\cite{ye2023m}, a similar trajectory of performance enhancement is also discerned, which exceeds the version with CLIP~\cite{radford2021learning} by 1.8\%.
It is important to note that the same datasets delineated in Sec.~\ref{sec:datasets} and experimental settings described in Sec.~\ref{sec:settings} are utilized to train these models to maintain a fair and consistent comparison.
Conclusively, our DoCo consistently enhances the visual representations of image encoder within LVLMs in text-rich scenarios, thereby bolstering the performance of visual document understanding.

Fig.~\ref{fig:qualitative_results} exhibits an assortment of qualitative outcomes engendered by Qwen-VL-Chat~\cite{bai2023qwen} between the version pre-trained with CLIP~\cite{radford2021learning} to the version optimized by DoCo across a diverse range of document images.
DoCo demonstrates proficiency in not only the extraction of relevant data from the document but also in providing corresponding responses by concentrating on unique regions.
The integration of contrastive learning, coupled with the multimodal features of document elements, enhances the efficacy of LVLMs, thereby facilitating a more thorough comprehension in scenarios abundant with text.
More visualizations are provided in the supplementary material.

\subsection{Ablation Study}
\begin{figure*}[t]
	\centering
	\includegraphics[width=1\linewidth]{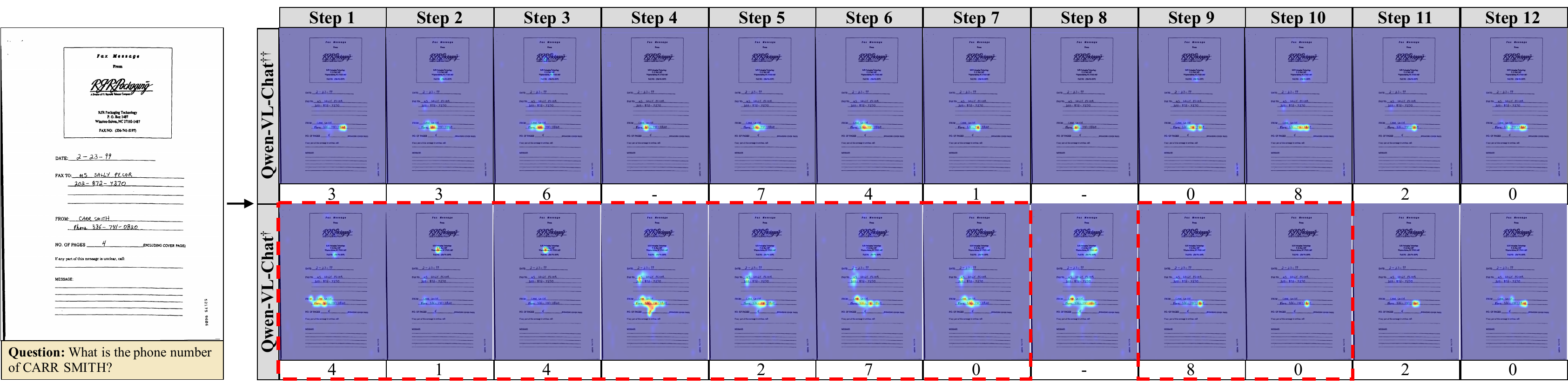}
	\caption{Visualization of the heat-maps and generated tokens in each step from CLIP~(``$\dag$'') and DoCo~(``$\dag\dag$'') based on Qwen-VL-Chat. CLIP yields attention maps exhibiting drift, which encounters the absence of fine-grained features. In contrast, our DoCo can accurately predict the attention position of the subsequent step with the context information, which demonstrates our method can assist the vision encoder in acquiring more effective visual cues in text-rich scenes. Best viewed in color.}
	\label{fig:analysis}
\end{figure*}
To investigate the efficacy of various components within our proposed DoCo, an extensive series of ablation studies based on the Qwen-VL-Chat~\cite{bai2023qwen} model are conducted on the DocVQA~\cite{mathew2021docvqa} dataset, which are summarized in Tab.~\ref{tab:ablation}.

\textbf{Effect of Intra-DoCo and Inter-DoCo:}
The data presented in Tab.~\ref{tab:ablation} substantiates the assertion that the integration of Intra-DoCo into Qwen-VL-Chat~(referred to as ``-w/\ Intra\&R\&I\&B\&T'') significantly enhances the performance by 1.7\% on the DocVQA dataset, in comparison to the version without DoCo involved in~(``-w/o DoCo''). 
This implies that the utilization of Intra-DoCo for pre-training LVLMs is more proficient for visual document understanding.
The observed performance enhancement can be ascribed to the congruence between patch-level image features and their corresponding multimodal contextual features, achieved through a fine-grained and localized scheme.
The implementation of Inter-DoCo into Qwen-VL-Chat~(referred to as ``$\dag\dag$'') further amplifies the average accuracy by 0.9\%, suggesting that the intricate interaction between images can assimilate contextual information, thereby helping the model discern textual details from a global standpoint.
These results further corroborate the efficacy of the proposed contrastive training pattern in enabling the vision encoder to capture more potent visual cues, thereby enhancing the comprehension of text-rich documents in LVLMs.

\begin{table}[t]
	\small
	\setlength{\tabcolsep}{0.1mm}
	\centering
	\begin{tabular}{l|cc|cc|ccc|c}
		\toprule[1.5pt]
		Method & Intra & Inter& R& A& I &B & T & DocVQA \\
		\hline
		Qwen-VL-Chat$^{*}{\rm_{w/o\ DoCo}}$ &\xmark&\xmark&\xmark& \xmark& \xmark  & \xmark & \xmark & 62.2\\
		Qwen-VL-Chat$^{*}{\rm_{w/\ Intra\&R\&I\&B\&T}}$  &\cmark&\xmark&\cmark& \xmark& \cmark& \cmark & \cmark & 63.9 \\
		Qwen-VL-Chat$^{*}{\rm_{w/\ DoCo\&A\&I\&B\&T}}$  &\cmark&\cmark&\xmark& \cmark& \cmark& \cmark & \cmark & 64.2 \\
		Qwen-VL-Chat$^{*}{\rm_{w/\ DoCo\&R\&T}}$  &\cmark&\cmark&\cmark& \xmark& \xmark& \xmark & \cmark & 63.7 \\
		Qwen-VL-Chat$^{*}{\rm_{w/\ DoCo\&R\&B\&T}}$  &\cmark&\cmark&\cmark& \xmark& \xmark& \cmark & \cmark & 64.4 \\
		\hline
		\textbf{Qwen-VL-Chat$^{\dag\dag}$} &\cmark&\cmark&\cmark& \xmark& \cmark  & \cmark & \cmark & \textbf{64.8} \\
		\bottomrule[1.5pt]
	\end{tabular}
	\caption{Ablation studies of DoCo based on Qwen-VL-Chat on DocVQA dataset. ``Intra'' and ``Inter'' denote ``Intra-DoCo'' and ``Inter-DoCo''. ``R'' and ``A'' stand for ``ROI Aggregation'' and ``Average Aggregation'' respectively. ``I'', ``B'' and ``T'' represent ``Image'', ``Box'' and ``Text'' modalities of document objects separately. ``w/o'' and ``w/'' are short for ``without'' and ``with''.}
	\label{tab:ablation}
\end{table}

\textbf{Effect of ROI Aggregation:}
To verify the capability of another core component ROI Aggregation, we juxtapose it with Average Aggregation, a technique that calculates the arithmetic mean of the patch features that overlap with the corresponding bounding boxes.
The empirical findings, as delineated in Tab.~\ref{tab:ablation}, underscore the superiority of ROI Aggregation~(referred to as ``$\dag\dag$'') over Average Aggregation~(referred to as ``-w/\ DoCo\&A\&I\&B\&T''), with an improvement of 0.6\%.
We hypothesize that the facilitation is derived from the inherent ability of ROI Aggregation to concentrate on the regions of interest, thereby collating more salient features for the text information.

\textbf{Effect of various modalities:}
Here, we aim to authenticate the contributions of various modalities towards the fine-grained features.
Compared with the performance of the model labeled as ``-w/o DoCo'', which is pre-trained via the traditional image-text contrastive learning at the whole image level akin to CLIP, the incorporation of the textual features of document objects optimized in the DoCo manner~(referred to as ``-w/\ DoCo\&R\&T'') results in an improvement of 1.5\% on the DocVQA dataset.
This confirms the benefit of alignment at the document objects level between visual features derived from the image encoder and the textual features from the multimodal encoder.
Encouragingly, we also discern that the introduction of the layout modality~(referred to as ``-w/\ DoCo\&R\&B\&T'') and image modality~(referred to as ``$\dag\dag$'') further augment performance by 0.7\% and 0.4\% respectively.
These observations demonstrate that the fine-grained representations furnished by the multimodal encoder can assist the image encoder of LVLMs in obtaining more effective cues for text-rich scenarios.

\subsection{Further Analysis on DoCo}
To provide intuitive examples for explaining the mechanism of our method, we visualize the heat-maps derived from the image encoder and the generated tokens at each decoding phase.
These models are pre-trained utilizing CLIP and DoCo, based on the Qwen-VL-Chat framework.
As depicted in Fig.~\ref{fig:analysis}, we surprisingly observe that the heat-maps generated by DoCo tend to consistently concentrate on the fine-grained textual features that have a contextual relationship, which proves that DoCo can assist the image encoder in capturing more efficacious visual cues in text-rich scenarios.
Upon the removal of DoCo, the activation maps produced by the original CLIP are scattered, a condition that is not favorable for the comprehension of visual documents.
Taking Fig.~\ref{fig:analysis} as a case study, the attention heat-maps of CLIP from the decoding steps 1 to 2, 3 to 4, and 8 to 9 exhibit significant drift and weak semantic interrelation, leading to a deviation from the correct response and yielding unsatisfactory outcomes.
These unstable behaviors can be ascribed to the absence of fine-grained features that are generated by the whole image-level discrimination.
By comparison, the model that has been pre-trained with DoCo is more adept at leveraging the context information to accurately predict the attention position of the subsequent step, which proves that aligning the multimodal features to the visual representations in the document object level is indeed a more preferable way to solve the fine-grained feature collapse problem.
These observations underscore the ability of DoCo to guide LVLMs in capturing salient details in a comprehensive manner, which is particularly crucial for tasks involving visual document understanding.

\section{Conclusion}
In this paper, we present a novel contrastive learning framework named Document Object COntrastive learning~(DoCo), which aims to extract fine-grained features for LVLMs in text-rich scenarios.
Different from previous image-level contrastive learning methods which discriminate the whole image instance between visual and textual inputs, DoCo discriminates the document objects between image and multimodal features within an image and across images, enabling the image encoder of LVLMs to learn more effective visual representations for text-rich scenes. 
Experiment results show that LVLMs equipped with DoCo can achieve superior performance and mitigate the gap between visual document understanding and conventional vision-language tasks.

\begin{appendices}
\section{Multimodal Feature Extraction}
\par \setlength{\parindent}{0em}
Within DoCo, we utilize LayoutLMv3~\cite{huang2022layoutlmv3} to extract the multimodal features of the document objects, which encompass textual embeddings, visual embeddings and layout embeddings.
This section provides a comprehensive description of the multimodal feature extraction process.

\textbf{Textual embeddings:} 
We pre-process the document images with an off-the-shelf OCR toolkit PaddleOCR~\cite{du2020pp} to obtain textual content and bounding boxes of the document objects.
The textual embeddings are subsequently extracted with the word embedding matrix from LayoutLMv3~\cite{huang2022layoutlmv3}, culminating in a sequence with a length of $T$ and a dimension of $d_{m}$. The maximum value of $T$ is established at 512.

\textbf{Visual embeddings:}
Inspired by LayoutLMv3~\cite{huang2022layoutlmv3}, the document image is resized to $H \times W$, and subsequently divided into a sequence of $P \times P$ patches.
These patches are then linearly projected to a dimension of $d_{m}$ and flattened into a sequence with a length of $I = HW / P^{2} = 196$.

\textbf{Layout embeddings:}
For the layout embeddings, we follow ~\cite{huang2022layoutlmv3} to involve 1D and 2D position embeddings to the $T$ textual tokens and $I$ image patches, where the 1D position refers to the index of tokens and the 2D position denotes the box coordinates of the corresponding object layouts.

As a result, these embeddings are forwarded to the multimodal encoder to aggregate the multimodal features $\mathbf{\widetilde{F}}^{m} = \left\lbrace \mathbf{f}^{m}_{1},  \mathbf{f}^{m}_{2}, ..., \mathbf{f}^{m}_{N+1}\right\rbrace \in \mathbb{R}^{(N + 1) \times d_{m}}$.
Note that each object is processed as a sequence of $\mathbf{f}^{m}_{i}$, and all the $N+1$ objects $\left\lbrace \mathbf{f}^{m}_{1},  \mathbf{f}^{m}_{2}, ..., \mathbf{f}^{m}_{N+1}\right\rbrace$ are processed in batch by the encoder.

\section{Fine-tuning Datasets}
During the fine-tuning stage, LVLMs undergo optimization utilizing approximately 0.4 million text-rich datasets, which include TextVQA~\cite{singh2019textvqa}, DocVQA~\cite{mathew2021docvqa}, ChartQA~\cite{masry2022chartqa}, OCRVQA~\cite{mishra2019ocr}, InfoVQA~\cite{mathew2022infographicvqa}, KLC~\cite{stanislawek2021kleister}, WTQ~\cite{pasupat2015wtq}, and TextCaps~\cite{sidorov2020textcaps}.
These datasets can be classified into three task categories: document Image captioning, key information extraction and document visual question answering.

\textbf{Document image captioning:}  Document image captioning involves generating descriptive text for a given document image, necessitating models to interpret and rationalize the text within these images to produce accurate captions. This process specifically requires models to integrate a novel modality of text present in the images, and to reason over both this text and the visual content within the image to generate comprehensive image descriptions. To enhance the performance of the model on the task of document image captioning, we utilize the training split of TextCaps~\cite{sidorov2020textcaps}.

\textbf{Key information extraction:} Key information extraction in document understanding, alternatively referred to as Property Extraction, denotes the methodological process of pinpointing and extracting salient or pertinent information from a specified document. This extracted information can span a wide array of data types, encompassing elements such as names, dates, geographical locations, organizational entities, financial figures, among other specific details that are integral to the comprehensive understanding of the document's content. In the present study, we leverage the training and validation splits of the KLC~\cite{stanislawek2021kleister} dataset to enhance the performance of our model in executing the key information extraction task.

\textbf{Document visual question answering:} Document visual question answering bears a superficial resemblance to knowledge information extraction in terms of structure. However, upon closer examination, the differences become more pronounced. Visual question answering necessitates the interpretation of an open-ended set of questions and the ability to handle a variety of document types, thereby requiring superior generalization capabilities. Moreover, the specific content under analysis necessitates a more profound understanding of visual elements, as the questions often pertain to figures and graphics that accompany the formatted text. To enhance the performance of our model, we utilize a variety of highly recognized public question-answering benchmark datasets, which include TextVQA~\cite{singh2019textvqa}, DocVQA~\cite{mathew2021docvqa}, ChartQA~\cite{masry2022chartqa}, OCRVQA~\cite{mishra2019ocr}, InfographicVQA~\cite{mathew2022infographicvqa} and WTQ~\cite{pasupat2015wtq}.

\section{More Interpretability on DoCo}
\begin{figure*}[htp]
	\centering
	\includegraphics[width=1\linewidth, height=22cm]{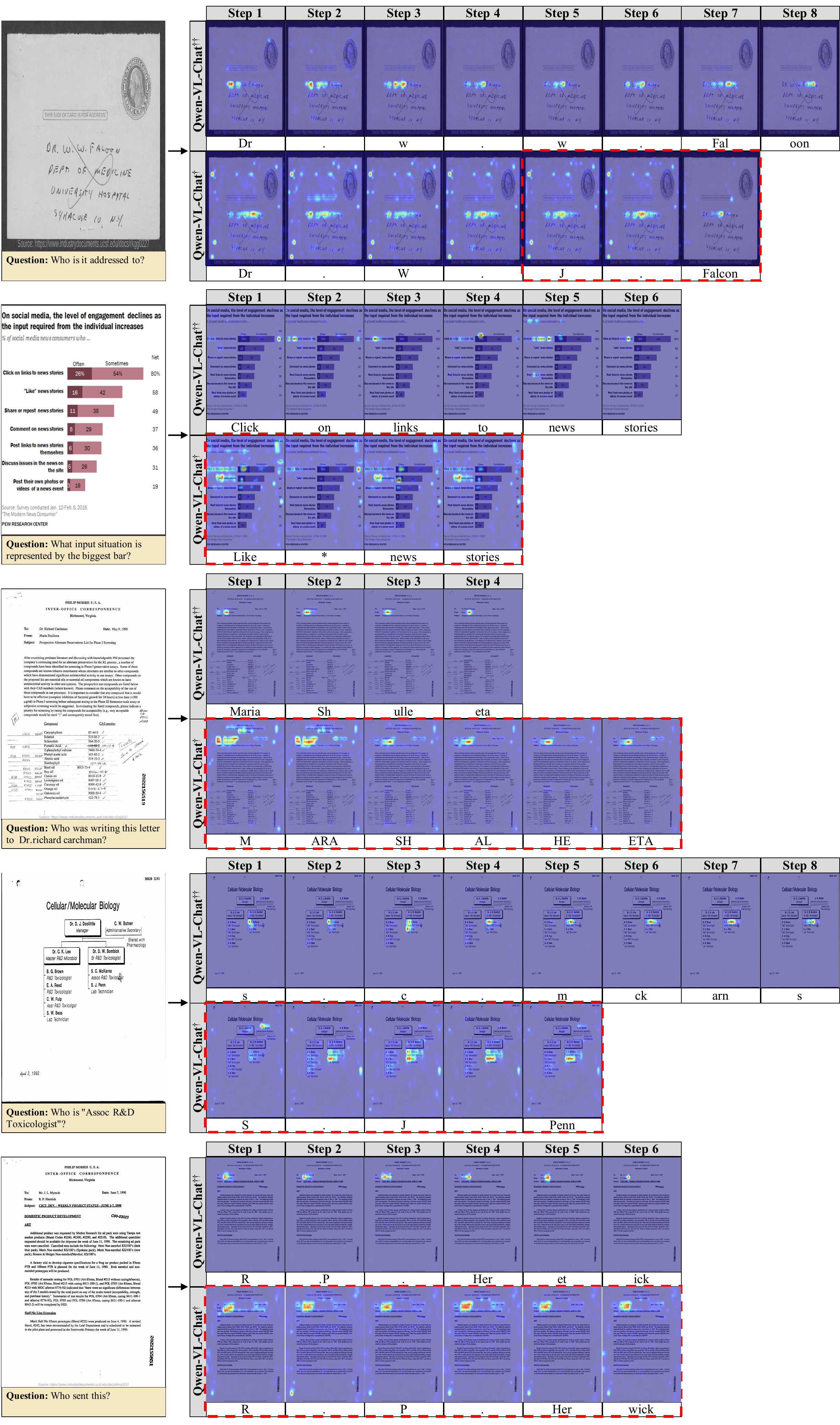}
	\caption{{Visualization of the heat-maps and generated tokens from CLIP~(``$\dag$'') and DoCo~(``$\dag\dag$'') in text-rich document scenarios.}} 
	\label{fig:app_vis}
\end{figure*}

\begin{figure*}[htp]
	\centering
	\includegraphics[width=1\linewidth, height=22cm]{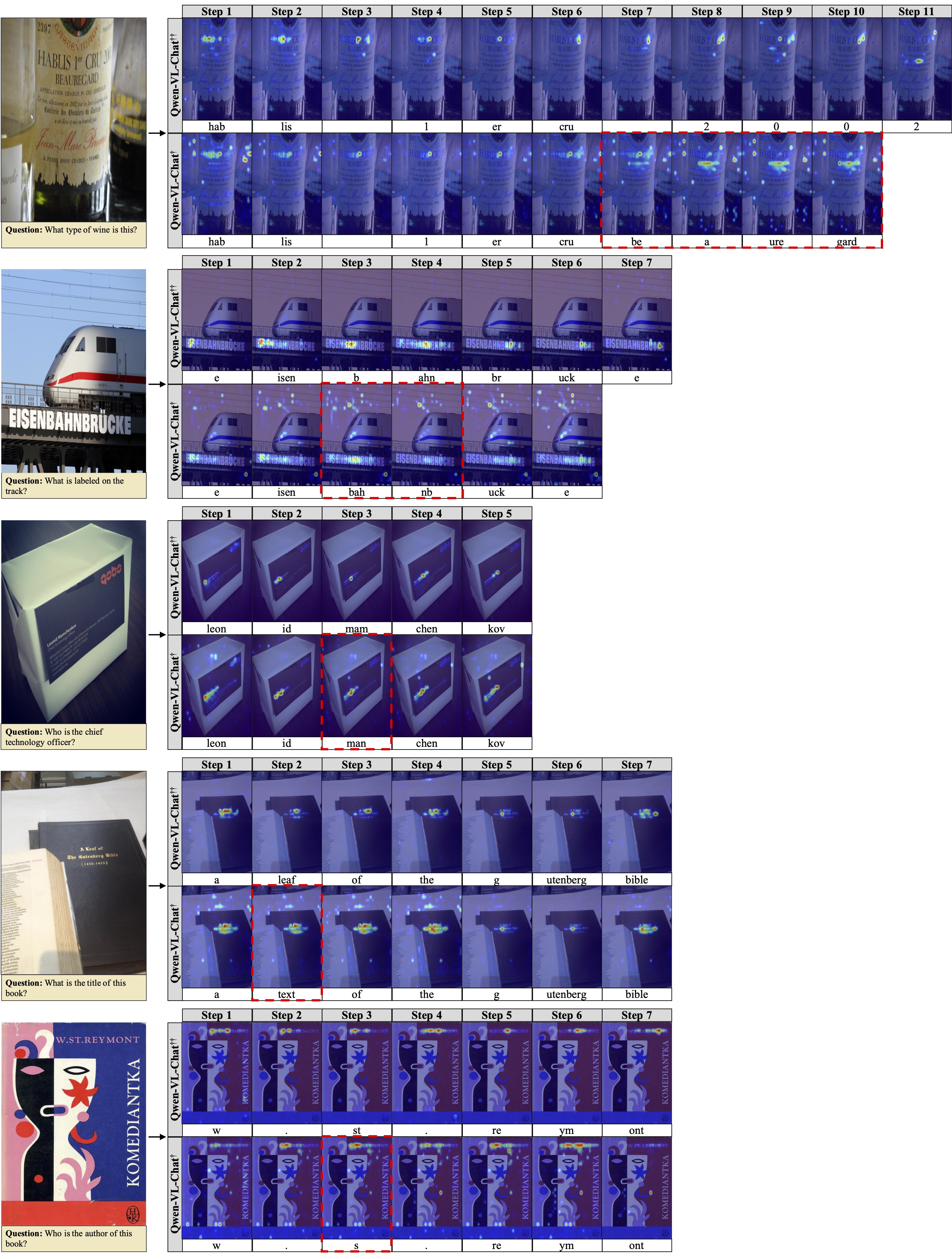}
	\caption{{Visualization of the heat-maps and generated tokens from CLIP~(``$\dag$'') and DoCo~(``$\dag\dag$'') in text-rich natural scenes.}} 
	\label{fig:app_vis2}
\end{figure*}

In Fig.~\ref{fig:app_vis} and Fig.~\ref{fig:app_vis2} of this appendix, we visualize more heat-maps derived from the image encoder and the generated tokens at each decoding phase between CLIP~\cite{radford2021learning} and DoCo based on the Qwen-VL-Chat~\cite{bai2023qwen} framework.

Fig.~\ref{fig:app_vis} elucidates the interpretability within text-rich document scenarios.
It is observed that the attention heat-maps of DoCo adeptly capture pertinent information associated with the correct response, whereas the maps of CLIP display a significant drift and weak semantic interrelation deviated from the ground truth, which is evident in the second and fourth instances of the figure.
Moreover, aligning the multimodal features to the visual representations in the document object level enhances comprehension of fine-grained text details and effectively mitigates recognition errors, which is corroborated by the first, third and fifth instances of the figure.

Fig.~\ref{fig:app_vis2} further illustrates additional results within text-rich natural scenes, following a similar pattern.
For instance, in the first case, the CLIP model generates incorrect tokens ``be-a-ure-gard'' from steps 7 to 10, which bear no relevance to the question ``type''.
Furthermore, the attention maps of CLIP display a significant lack of focus and drift from step 7, resulting in the absence of fine-grained features.
By comparison, the maps of DoCo tend to focus on the continuous fine-grained textual features, yielding satisfactory results.
Additionally, our DoCo leverages the context information to predict the subsequent results, as demonstrated by the image features of ``2002'' in step 11, which are fuzzy but can be inferred to the correct tokens from the context ``en 2002 par'' below.

In summary, Fig.~\ref{fig:app_vis} and Fig.~\ref{fig:app_vis2} compellingly demonstrate that DoCo can assist the image encoder in capturing more effective visual cues in text-rich scenarios.

\section{More Qualitative Results}
\begin{figure*}[htp]
	\centering
	\includegraphics[width=1\linewidth]{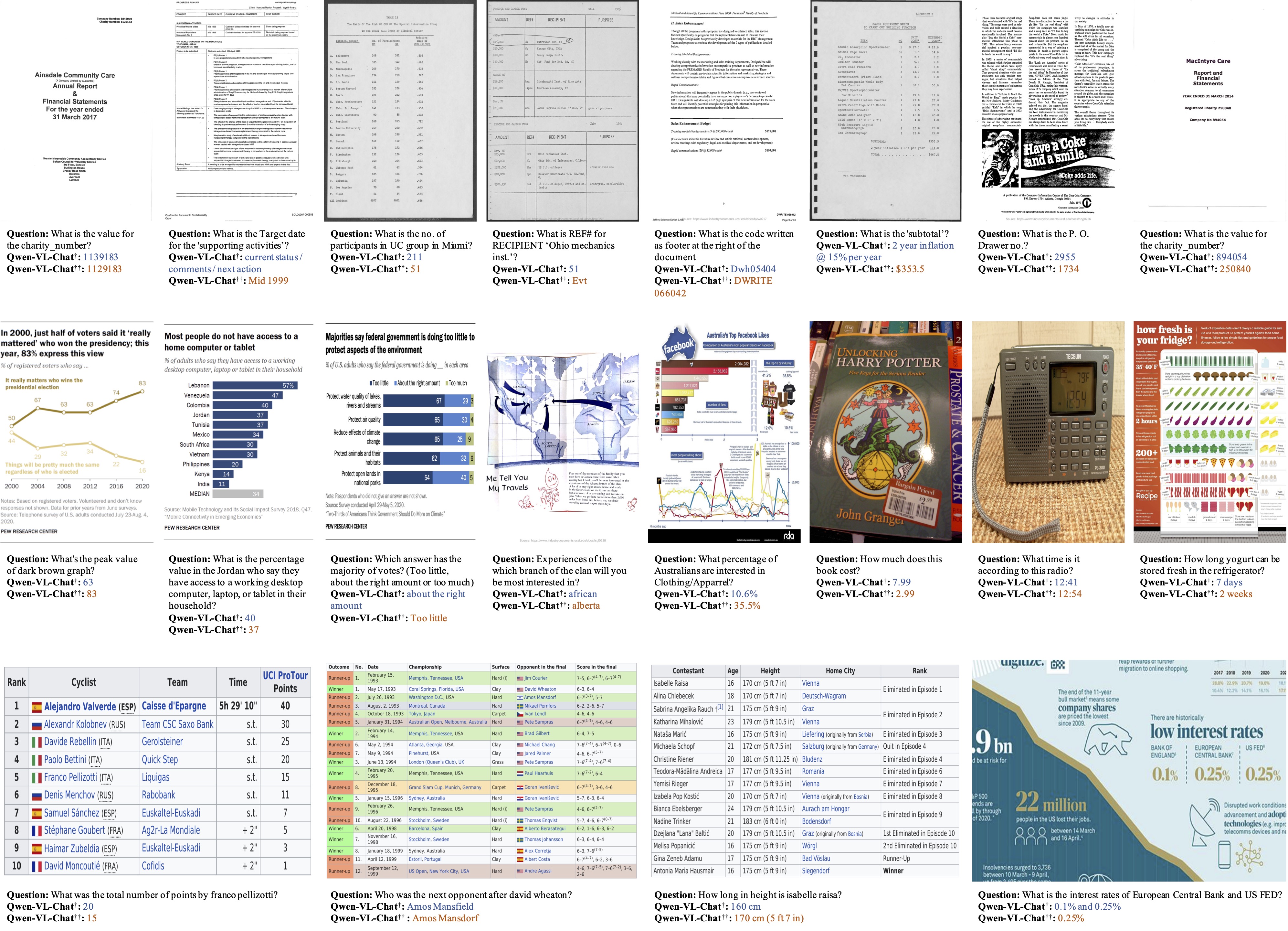}
	\caption{{More qualitative results between CLIP~(``$\dag$'') and DoCo~(``$\dag\dag$'') based on the Qwen-VL-Chat model in text-rich scenes.}} 
	\label{fig:app_goodcase}
\end{figure*}

Fig.~\ref{fig:app_goodcase} presents an expanded set of qualitative outcomes derived from a variety of benchmark datasets between CLIP and DoCo based on Qwen-VL-Chat~\cite{bai2023qwen}.
The illustrations underscore the superior generalization capacity of DoCo, which assists the vision encoder of LVLMs in acquiring more efficacious cues and enhances comprehension in text-rich scenes.

\begin{figure*}[htp]
	\centering
	\includegraphics[width=1\linewidth]{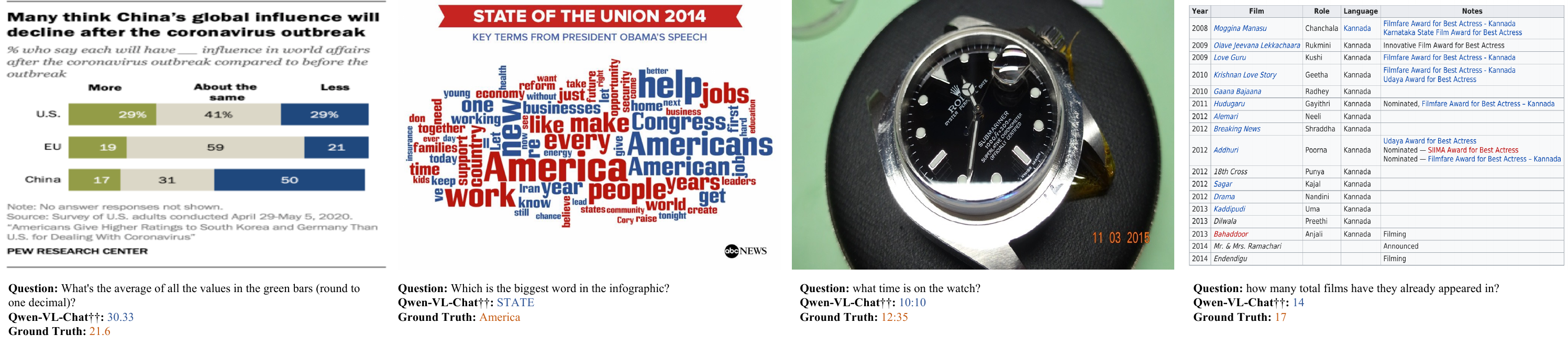}
	\caption{{Failure cases of DoCo on document-related commonsense reasoning and mathematical calculations.}} 
	\label{fig:app_badcase}
\end{figure*}

We also delineate the failure cases of our method in Fig.~\ref{fig:app_badcase}.
It is evident that our DoCo still struggles with document-related commonsense reasoning and mathematical calculations, which furnishes invaluable insights for the enhancement of document comprehension capabilities with LVLMs in this domain. 
Future research endeavors will investigate these problems and attempt to attack them by further improving visual understanding performance.

\section{Broader Impact}
The remarkable proficiencies of LVLMs hold vast potential for facilitating more robust document analysis and comprehension in text-rich environments, but the significance of fine-grained features remains largely unexplored within the LVLM community.
Thanks to the document object discrimination between visual and multimodal representations, our proposed DoCo tailored for the fine-grained feature collapse issue can yield more precise results in text-rich scenarios.
The acquisition of more efficient fine-grained visual representations opens up a plethora of potential applications and opportunities for visual document understanding tasks.
We advocate for researchers to develop LVLMs integrated with DoCo for text-rich tasks, as we anticipate this to be particularly advantageous.


\end{appendices}
{
    \small
    \bibliographystyle{ieeenat_fullname}
    \bibliography{main}
}

\end{document}